\documentclass[twocolumn,letterpaper]{article}
\usepackage{graphicx} 
\usepackage{amsmath}
\usepackage{bbding}
\usepackage{multirow}
\usepackage{makecell}
\usepackage{colortbl}
\usepackage{caption}
\usepackage{authblk} 
\definecolor{zbbest}{rgb}{0.96, 0.57, 0.58}
\definecolor{zbsecond}{rgb}{0.98, 0.78, 0.57}
\definecolor{zbthird}{rgb}{1.0, 1.0, 0.56}

\definecolor{gscolor}{rgb}{1.0,0.6,0.0} %

\usepackage[pagebackref,breaklinks,colorlinks, citecolor=yellow]{hyperref}

\title{A Refined 3D Gaussian Representation for High-Quality Dynamic Scene Reconstruction}
\author{
    Bin Zhang, 
    Bi Zeng$^{\dag}$
    and Zexin Peng \\
    Guangdong University of Technology, Guangzhou, Guangdong, People’s Republic of China \\
    $^{\dag}$Corresponding author \\
    \url{zomb2000@163.com}
}

\begin{document}


\twocolumn[ {
\maketitle
\renewcommand\twocolumn[1][]{#1}%
\begin{center}
    \centering
    \captionsetup{type=figure}
    \includegraphics[width=1.0\linewidth]{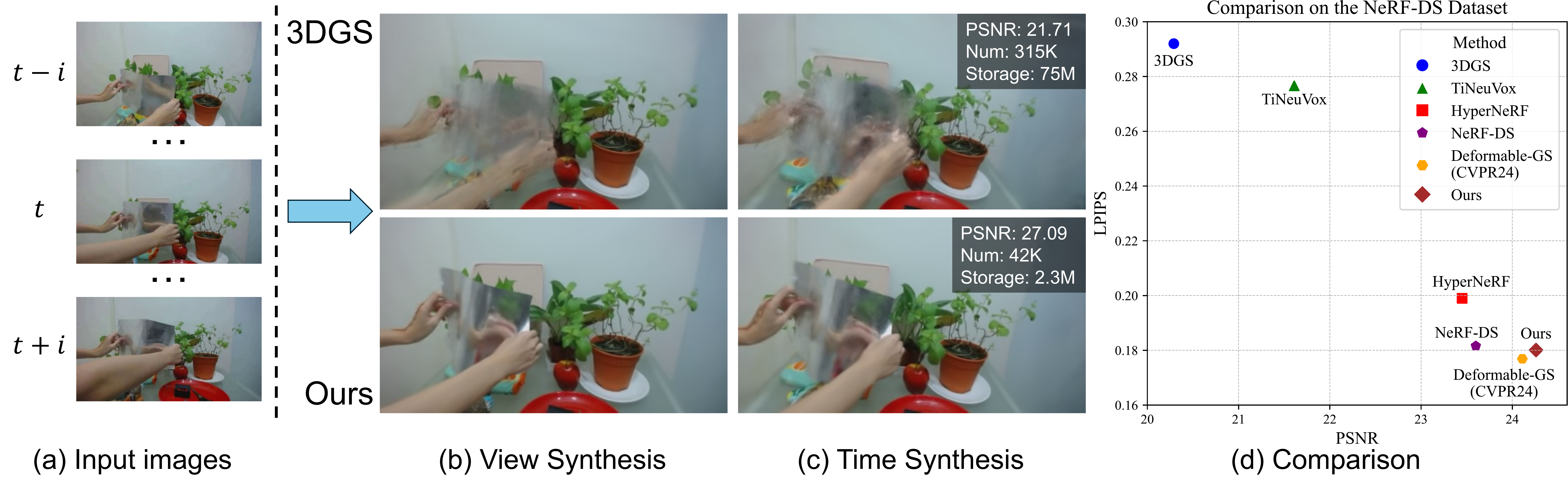}
    \captionof{figure}{Given a set of monocular multi-view images and camera poses (a), our approach not only facilitates the rendering of dynamic scenes (b) but also enables novel view time interpolation (c), leading to superior rendering quality compared to existing methods (d), alongside a notable reduction in memory usage.}
\label{fig:over_show}
\end{center}%
} ]

\begin{abstract}
In recent years, Neural Radiance Fields (NeRF) has revolutionized three-dimensional (3D) reconstruction with its implicit representation. Building upon NeRF, 3D Gaussian Splatting (3D-GS) has departed from the implicit representation of neural networks and instead directly represents scenes as point clouds with Gaussian-shaped distributions. While this shift has notably elevated the rendering quality and speed of radiance fields but inevitably led to a significant increase in memory usage. Additionally, effectively rendering dynamic scenes in 3D-GS has emerged as a pressing challenge. To address these concerns, this paper purposes a refined 3D Gaussian representation for high-quality dynamic scene reconstruction. Firstly, we use a deformable multi-layer perceptron (MLP) network to capture the dynamic offset of Gaussian points and express the color features of points through hash encoding and a tiny MLP to reduce storage requirements. Subsequently, we introduce a learnable denoising mask coupled with denoising loss to eliminate noise points from the scene, thereby further compressing 3D Gaussian model. Finally, motion noise of points is mitigated through static constraints and motion consistency constraints. Experimental results demonstrate that our method surpasses existing approaches in rendering quality and speed, while significantly reducing the memory usage associated with 3D-GS, making it highly suitable for various tasks such as novel view synthesis, and dynamic mapping.

\end{abstract}

\section{Introduction}

For various applications including augmented reality (AR), virtual reality (VR) and 3D content production, the accurate reconstruction and rendering of dynamic scenes from a set of input images with high fidelity and realism are imperative. As described in ~\cite{colletHighqualityStreamableFreeviewpoint2015, kanadeVirtualizedRealityConstructing1997, liTemporallyCoherentCompletion2012}, conventional approaches employed for modeling these dynamic scenes predominantly leaned on surface-based representations. Nonetheless, implicit scene representations such as NeRF~\cite{mildenhallNeRFRepresentingScenes2021} have showcased exceptional prowess in tasks such as novel view synthesis and scene reconstruction.

NeRF not only achieves efficient rendering of 3D scenes but also significantly reduces the storage space required for the scenes. By integrating deformation fields~\cite{pumarolaDNeRFNeuralRadiance2021} or adding a temporal dimension, and combining with tensor decomposition techniques~\cite{shaoTensor4DEfficientNeural2023}, dynamic scene modeling can be satisfactorily accomplished. Nevertheless, NeRF is hindered by prolonged training durations and slow rendering speeds. Despite notable advancements such as Instant-ngp~\cite{mullerInstantNeuralGraphics2022} and Plenoxels~\cite{fridovich-keilPlenoxelsRadianceFields2022}, which reduces training time and improves rendering speeds through the fusion of explicit grids with implicit features, they still fail to meet real-time requirements.
  
Subsequently, 3D-GS~\cite{kerbl3DGaussianSplatting2023} emerged as a novel solution, diverging from the implicit representation of neural networks to directly represented scenes as point clouds with Gaussian-shaped distributions. This fully explicit representation led to a rendering speed improvement of NeRF by more than a $100\times$ speedup and significantly enhanced rendering quality. However, this will also lead to a significant increase in storage usage~\cite{leeCompact3DGaussian2023}. Furthermore, rendering dynamic scenes within 3D-GS has become a new challenge. Despite recent methods such as Deformable-GS~\cite{yangDeformable3DGaussians2023a} and 4D-GS~\cite{wu4DGaussianSplatting2023} aimed at tackling the rendering issues of dynamic scenes in 3D-GS, they have yet to resolve the problem of memory consumption. Moreover, using neural networks to represent the dynamic offset of points also reduces rendering speed~\cite{yangDeformable3DGaussians2023a}.

In this context, this paper combines the advantages of NeRF and 3D-GS, introducing a dynamic scene rendering framework based on a hybrid representation of explicit and implicit features. While addressing the representation of dynamic scenes, our framework aims to mitigate the storage consumption associated with 3D-GS and introduces a novel view synthesis method tailored for dynamic mapping.

Firstly, we employ deformation fields to represent the dynamic offset of Gaussian points, alongside leveraging hash encoding and a tiny MLP to express the color features of 3D-GS points, thus reducing their storage consumption. Furthermore, we introduce a learnable denoising mask to filter out noise points from the scene, further compressing 3D-GS. Finally, we utilize static constraints and motion consistency constraints to mitigate motion artifacts in the points, ensuring the accuracy and efficiency of the rendering framework when processing dynamic scenes.

In summary, the major contributions of our work are as follows:
\begin{itemize}
    \item A hybrid representation is adopted by combining deformation fields, hash encoding, and 3D-GS, which significantly reduces memory usage while achieving efficient and realistic rendering of dynamic scenes.
    \item A learnable denoising mask in conjunction with noise loss is proposed, which can effectively identify and remove noise points in 3D-GS, thereby enhancing the purity and quality of scene rendering.
    \item Static constraints and motion consistency constraints are introduced to minimize noise in points during motion, ensuring that deformation fields can accurately learn the dynamic offset of points.
\end{itemize}
\section{Related Work}

\subsection{Dynamic NeRF}
In recent years, synthesizing novel views of dynamic scenes has emerged as a key focus within the academic community. Initially, NeRF~\cite{mildenhallNeRFRepresentingScenes2021} employed MLPs to implicitly model static scenes, while subsequent research~\cite{guoForwardFlowNovel2023b,liNeuralSceneFlow2021,pumarolaDNeRFNeuralRadiance2021}, expanded its scope to encompass dynamic scenes by introducing deformation fields. For instance, HyperNeRF~\cite{parkHyperNeRFHigherdimensionalRepresentation2021a} learns a hyperspace to represent bases for typical shapes across multiple dimensions.~\cite{gaoDynamicViewSynthesis2021, liNeural3DVideo2022a} model dynamic scenes as 4D radiance fields, but due to the demands of ray-point sampling and volume rendering, this leads to extremely high computational costs. To address this issue, Fang \textit{et al.}~\cite{fangFastDynamicRadiance2022} first utilized voxel grid representation to train dynamic NeRFs, achieving training speeds far exceeding those of neural network-based methods. Guo \textit{et al.}~\cite{guoNeuralDeformableVoxel2022} also proposed a deformable voxel grid method for rapid training. Other methods based on grid or plane structures~\cite{caoHexPlaneFastRepresentation2023,fridovich-keilKPlanesExplicitRadiance2023,shaoTensor4DEfficientNeural2023,wangMixedNeuralVoxels2023,wangNeuralResidualRadiance2023,parkTemporalInterpolationAll2023} have improved speed and performance in various dynamic scenes. However, these hybrid models have certain limitations in expressing high-frequency information.

\subsection{Dynamic Gaussian Splatting}
On the other hand, 3D-GS~\cite{kerbl3DGaussianSplatting2023}, with its fully explicit representation, is adept at conveying higher-frequency information, and certain works have successfully utilized it to represent dynamic scenes. 4D Gaussian Splatting~\cite{luScaffoldGSStructured3D2023, leeDeblurring3DGaussian2024, malarzGaussianSplattingNeRFbased2024} extends the concept of 3D-GS to dynamic scenes by introducing a temporal dimension. This extension enables the representation and rendering of scenes that evolve over time, significantly enhancing the real-time rendering performance of dynamic scenes while upholding high-quality visual output. For instance, Scaffold-GS~\cite{luScaffoldGSStructured3D2023} focus on applying 3D-GS to dynamic scene modeling, introducing techniques for tracking and representing moving scene elements. Additionally, 4D-GS~\cite{wu4DGaussianSplatting2023} combines neural voxels to achieve dynamic scene reconstruction.~\cite{luitenDynamic3DGaussians2023} not only synthesizes novel views of dynamic scenes but also accomplishes 6 DOF tracking of all dense scene elements. Works represented by Deformable-GS~\cite{yangDeformable3DGaussians2023a, huangSCGSSparseControlledGaussian2023,liangGauFReGaussianDeformation2023,shawSWAGSSamplingWindows2023} introduce the concept of deformation fields into 3D-GS. Among them, SC-GS~\cite{huangSCGSSparseControlledGaussian2023} achieves state-of-the-art (SOTA) performance on the D-NeRF\cite{pumarolaDNeRFNeuralRadiance2021} dataset by combining sparse control points with Deformable MLP. GauFRe~\cite{liangGauFReGaussianDeformation2023} explores novel applications and editing functionalities in dynamic scenes.

\begin{figure*}
    \centering
    \includegraphics[width=\linewidth]{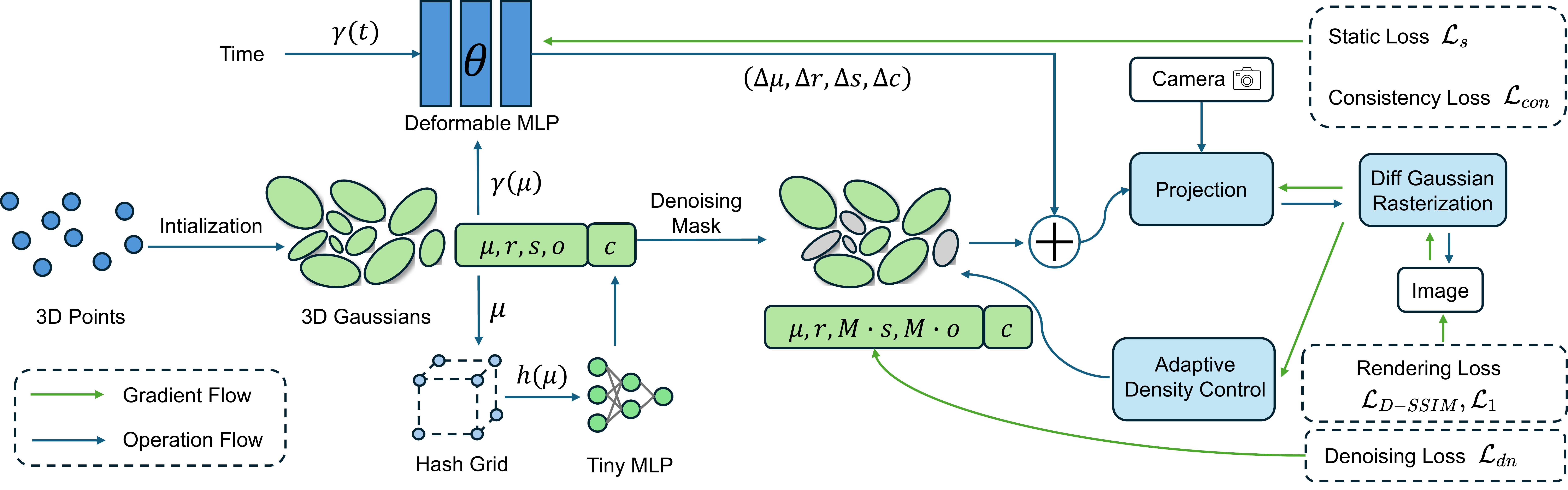}
    \caption{\textbf{Overview of our pipeline.} The details of the overall framework are elaborated in Section~\ref{sec:overview}. How deformation fields and hash encoding collaborate for dynamic representation is elaborated in Section~\ref{sec:deformable_3D_hash}. The denoising mask is introduced in Section~\ref{sec:denoising_mask}. Finally, in Section~\ref{sec:static_consistency}, we provide a concrete implementation of static constraints and motion consistency constraints, demonstrating how they facilitate the network in better learning the dynamic offsets of points in the scene.}
    \label{fig:overview}
\end{figure*}

\subsection{Compact Gaussian Splatting}
There are efforts dedicated to compressing and pruning Gaussian points to optimize the representation and rendering efficiency of scenes. LightGaussian~\cite{fanLightGaussianUnbounded3D2024} reduces unnecessary computational burden by pruning Gaussian points that contribute less to scene reconstruction. Lee \textit{et al.}~\cite{leeCompact3DGaussian2023} introduces a novel volume-based masking strategy that effectively reduces the number of Gaussian points required to represent scenes without compromising rendering performance. Niedermayr \textit{et al.}~\cite{niedermayrCompressed3DGaussian2024} compress both color and Gaussian parameters into a compact codebook, employing sensitivity measures for efficient quantization and fine-tuning, and utilizing entropy coding to leverage spatial consistency.

\section{Preliminaries}
\label{sec:preliminaries}
\par 3D Gaussian splatting~\cite{kerbl3DGaussianSplatting2023} represents an advancing graphics technique that conceptualizes a scene as comprising point clouds following Gaussian distributions. Each Gaussian point is centered at a 3D coordinate $\mu$ and possesses a 3D variance matrix $\Sigma$ to characterize its shape distribution, expressed specifically as:
\begin{equation}
    Ga(x)=e^{-\frac12(x-\mu)^T\Sigma^{-1}(x-\mu)}.
\end{equation}
Here, the variance matrix $\Sigma$ is defined as $\Sigma=RSS^TR^T$, where $R \in SO(3)$ represents a rotation matrix, initially expressed in terms of quaternions $r$ during the optimization process, and $S$ denotes a scale matrix expressed utilizing a vector $s$. Both $r$ and $s$ undergo optimization in the actual optimization procedure. Additionally, the transformation of the 3D variance matrix of Gaussian points into the variance matrix of pixels on a 2D image is accomplished through the following expression:
\begin{equation}
    \Sigma'=JW \Sigma W^TJ^T,
\end{equation}
where $J$ denotes the Jacobian matrix of the projection matrix, while $W$ represents the matrix for camera view transformation. Additionally, each Gaussian point is associated with color $c$ and opacity $o$. When combined with $\Sigma$, these parameters facilitate the computation of $\alpha$ for each point during volume rendering:
\begin{equation}
    \alpha_i=\sigma_ie^{-\frac12(p-\mu_i^{\prime})^T\Sigma_i^{\prime}(p-\mu_i^{\prime})},
\end{equation}
where $\mu^{'}$ denotes the center coordinate of the Gaussian point projected onto the 2D image. Similar to volume rendering in NeRF~\cite{mildenhallNeRFRepresentingScenes2021}, 3D-GS determines the RGB value of each pixel by aggregating the weighted $\alpha$ and $c$  of each point.
\begin{equation}
    C=\sum_{i=1}^NT_i\alpha_i\mathbf{c}_i, \mathrm{where~}T_i=\Pi_{j=1}^{i-1}(1-\alpha_j).
\end{equation}
Finally, through the optimization of parameters for all Gaussian points $\{G_i:{\mu_i, r_i, s_i, c_i, o_i}\}$, coupled with the implementation of an adaptive densification strategy for Gaussian points, high-fidelity image rendering can be attained.

\section{Method}
\subsection{Overview}
\label{sec:overview}
\par Firstly, the optimization process commences with Structure from Motion (SfM) points derived from COLMAP or generated randomly, serving as the initial state for the 3D Gaussians. We use position encoding $\gamma(\mu)$ and time encoding $\gamma(t)$ of 3D Gaussians as inputs, passed to a deformable MLP network to predict the offset $(\Delta \mu, \Delta r, \Delta s, \Delta c)$ of dynamic 3D Gaussians in the canonical space. For color representation, we employ hash encoding combined with a tiny MLP. Concurrently, we utilize a denoising mask to mask the opacity $o$ and scale $s$ of points, which, upon addition with the offset values, yields ${G: \{\mu+\Delta \mu, r+\Delta r, M \cdot s+ \text{sg}(M) \cdot \Delta s, c+\Delta c, M \cdot o\}}$. Finally, static constraints and motion consistency constraints ensure that the deformation field accurately learns the dynamic offset of points. Moreover, the rendering process and point densification align with 3D-GS, as detailed in Section~\ref{sec:preliminaries}.

\subsection{Deformable 3D Hash Gaussian}
\label{sec:deformable_3D_hash}
\par In our approach to representing dynamic scenes, we integrate the deformation field with hash encoding to formulate a distinct model structure, illustrated in the figure~\ref{fig:overview}. The concept of the deformation field is derived from D-NeRF~\cite{pumarolaDNeRFNeuralRadiance2021}, wherein the coordinate of 3D points, along with their temporal attributes, are encoded within an $8 \times 256$ MLP network. This process yields the coordinate offsets for these points. By employing neural networks, this strategy adeptly learns temporal features, facilitating accurate simulation of the dynamic fluctuations in sampled point positions over time.
  
\par The encoded features of Gaussian point spatial coordinates $\mu$ and time $t$ serve as inputs, yielding coordinate offsets $\Delta \mu$, quaternion offsets $\Delta r$, scale offsets $\Delta s$, and color offsets $\Delta c$.
\begin{equation}
    (\Delta \mu,\Delta r,\Delta s, \Delta c)=\mathcal{F}_\theta(\gamma(\mu),\gamma(t)),
\end{equation}
where $\mathcal{F}_\theta$ denotes the 8x256 deformation field MLP, and $\gamma(\cdot)$ represents the frequency-position encoding, expressed as follows:
\begin{equation}
    \gamma(p)=(sin(2^k\pi p),cos(2^k\pi p))_{k=0}^{L-1},    
\end{equation}
where $L$ denotes the order of frequency encoding.

\par Reflecting on the original 3D-GS model, the color attributes of each Gaussian point are depicted using optimizable third-order spherical harmonics functions. It is applied individually to each of the RGB channels, with each channel consisting of 16 independent 32-bit floating-point variables, collectively defining the color representation of each Gaussian point. Consequently, color information occupies up to 81\% of the total storage. During training, to ensure the optimization effectiveness of all Gaussian points, it is necessary to retain the spherical harmonics values corresponding to each point. This imposes exceedingly high demands on storage resources and optimization efficiency.

To tackle this challenge, we drew inspiration from in Instant-ngp~\cite{mullerInstantNeuralGraphics2022} and incorporated hash encoding techniques to encode, along with a tiny MLP for decoding, the color characteristics of Gaussian points. Experimental findings confirmed that this approach not only adeptly captures the color features of Gaussian points but also notably mitigates storage demands and optimization burdens. To attain convergence in dynamic scene between the deformation field and hash expression, we directly add the color $c$ output by the tiny MLP to the offset $\Delta c$ yielded by the deformable MLP.:
\begin{equation}
    c_i = \Phi_\theta (h(\mu_i)) + \Delta c_i,
\end{equation}
where $h(\cdot)$ denotes the hash grid feature encoding, and $\Phi_\theta$ represents the tiny MLP, which is $64 \times 2$, serving as a feature decoder. During the actual optimization process, this approach not only achieves rapid convergence but also reduces the storage requirement to 20\% of its original size. The final expression for the deformable 3D hash Gaussian field is as follows: 
\begin{equation}
    {G_i:\{\mu_i+\Delta \mu_i, r_i+\Delta r_i, s_i+\Delta s_i, c_i+\Delta c_i, o_i\}}.
\end{equation}

\subsection{Learnable Denoising Mask}
\label{sec:denoising_mask}
In the original rendering process, Gaussian points undergo simple cloning and splitting based on the gradient changes of Gaussian point coordinates $\mu$ during the optimization process. This mechanism achieves effective point cloud densification while filtering out points with low contribution, i.e., opacity $o$, lower than a preset threshold. Although demonstrating simplicity and efficiency in point densification and ineffective point removal, this method still struggles to avoid generating a certain number of noise points which may introduce artifacts in the rendering results. This not only deteriorates the rendering quality but also results in excessive storage and loss of computational resources. To tackle this issue, we propose a learnable mask $M_i$ and combine it with the deformation field to eliminate noise points in dynamic scenes:
\begin{equation}
    M_{i}=\sigma(m_{i}),
\end{equation}
where $\sigma(\cdot)$ denotes the sigmoid function, which maps the learnable parameter $m_n$ to  $(0,1)$. During optimization, this learnable mask effectively eliminates some noise points and notably enhances rendering quality. However, after referring to~\cite{leeCompact3DGaussian2023},  which shares similarities with our work, we adjusted the learnable mask to the following expression:
\begin{equation} \label{eq:mask}
    M_i=\sigma(m_i)+\text{sg}(1[\sigma(m_i)>\epsilon]-\sigma(m_i)),
\end{equation}
where $\text{sg}(\cdot)$ denotes the operation of stopping gradient computation preventing the gradients of certain tensors from being calculated during backpropagation. $\epsilon$ represents an artificially set threshold. Through equation~\eqref{eq:mask}, we transform the continuous mask $M_i =\sigma(m_i)$ into a binary mask, which can remove more ineffective points, although potentially leading to a slight decrease in rendering quality. By integrating the scale offset $\Delta s$, we utilize the binary mask to mask the opacity $o$ and scale vector $s$:
\begin{equation}
    s_i = M_i \cdot s_i + \text{sg}(M_i)\cdot\Delta s_i, o_i=M_i \cdot o_i.
\end{equation}
To enhance the effectiveness of training this mask, we introduce a denoising constraint. We establish expected values $E(o)$ for opacity $o$ and $E(s)$ for the scale $s$. To enable the mask to learn how to identify and remove noise points, we aim to minimize the discrepancy between the expected values and the actual values, thereby bringing the actual values closer to the expected ones. Specifically, our objective is to minimize the following error:
\begin{equation}
\begin{split}
    \mathcal{L}_{dn} = & \frac1N\sum_{i=1}^N\| E(s_i)- (M_i \cdot s_i +\text{sg}(M_i)\cdot \Delta s_i) \|\\ 
    & + \frac1N\sum_{i=1}^N\| E(o_i)-M_i \cdot o_i \|.
\end{split}
\end{equation}
Here, the expected values $E(\cdot)$ are derived by computing the average within a historical sliding window throughout the training process. During the actual rendering procedure, we incorporate this loss function starting from the 5,000th iteration, enabling it to contribute to the optimization process. This methodology encourages the mask to more adeptly identify and mask noisy points. In addition to enabling the denoising mask to remove more points, we incorporate the following constraints:
\begin{equation}
    \mathcal{L}_{m}=\frac1N\sum_{n=1}^N\sigma(m_n).
\end{equation}

\subsection{Static and Consistency Constraints}

\label{sec:static_consistency}
\begin {table*}[ht]
\centering
\resizebox{\textwidth}{!}{
\begin{tabular}{ccccccccccccccccc}
\hline \multirow[b]{2}{*}{ Method } & \multicolumn{3}{c}{ Sieve } & \multicolumn{3}{c}{ Plate } & \multicolumn{3}{c}{ Bell } & \multicolumn{3}{c}{ Press } \\
& PSNR$\uparrow$ & SSIM$\uparrow$ & LPIPS$\downarrow$ & PSNR$\uparrow$ & SSIM$\uparrow$ & LPIPS$\downarrow$ & PSNR$\uparrow$ & SSIM$\uparrow$ & LPIPS$\downarrow$ & PSNR$\uparrow$ & SSIM $\uparrow$ & LPIPS$\downarrow$ \\

\hline 3D-GS~\cite{kerbl3DGaussianSplatting2023} & 23.16 & 0.8203 & 0.2247 & 16.14 & 0.6970 & 0.4093 & 21.01 & 0.7885 & 0.2503 & 22.89 & 0.8163 & 0.2904 \\

TiNeuVox~\cite{fangFastDynamicRadiance2022} & 21.49 & 0.8265 & 0.3176 & \cellcolor{zbsecond}20.58 & 0.8027 & 0.3317 & 23.08 & \cellcolor{zbthird}0.8242 & 0.2568 & 24.47 & 0.8613 & 0.3001 \\

HyperNeRF~\cite{parkHyperNeRFHigherdimensionalRepresentation2021a} & 25.43 & \cellcolor{zbthird}0.8798 & 0.1645 & 18.93 & 0.7709 & 0.2940 & 23.06 & 0.8097 & 0.2052 & \cellcolor{zbsecond}26.15 & \cellcolor{zbbest}0.8897 & \cellcolor{zbthird}0.1959 \\

NeRF-DS~\cite{yanNeRFDSNeuralRadiance2023} & \cellcolor{zbsecond}25.78 & \cellcolor{zbbest}0.8900 & \cellcolor{zbbest}0.1472 & \cellcolor{zbthird}20.54 & \cellcolor{zbthird}0.8042 & \cellcolor{zbbest}0.1996 & \cellcolor{zbthird}23.19 & 0.8212 & \cellcolor{zbthird}0.1867 & 25.72 & 0.8618 & 0.2047 \\

Deformable-GS~\cite{yangDeformable3DGaussians2023a} & \cellcolor{zbthird}25.70 & 0.8715 & \cellcolor{zbsecond}0.1504 & 20.48 & \cellcolor{zbsecond}0.8124 & \cellcolor{zbsecond}0.2224 & \cellcolor{zbbest}25.74 & \cellcolor{zbbest}0.8503 & \cellcolor{zbbest}0.1537 & \cellcolor{zbthird}26.01 & \cellcolor{zbsecond}0.8646 & \cellcolor{zbbest}0.1905 \\ 

Ours & \cellcolor{zbbest}26.31 & \cellcolor{zbsecond}0.8819 & \cellcolor{zbthird}0.1547 & \cellcolor{zbbest}20.59 & \cellcolor{zbbest}0.8141 & \cellcolor{zbthird}0.228 & \cellcolor{zbsecond}25.03 & \cellcolor{zbsecond}0.8492 & \cellcolor{zbsecond}0.1567 & \cellcolor{zbbest}26.42 & \cellcolor{zbthird}0.8624 & \cellcolor{zbsecond}0.1957 \\

\hline & \multicolumn{3}{c}{ Cup } & \multicolumn{3}{c}{ As } & \multicolumn{3}{c}{ Basin } & \multicolumn{3}{c}{ Mean } \\

Method & PSNR$\uparrow$ & SSIM$\uparrow$ & LPIPS$\downarrow$ & PSNR$\uparrow$ & SSIM$\uparrow$ & LPIPS$\downarrow$ & PSNR$\uparrow$ & SSIM$\uparrow$ & LPIPS$\downarrow$ & PSNR$\uparrow$ & SSIM $\uparrow$ & LPIPS$\downarrow$ \\

\hline 3D-GS~\cite{kerbl3DGaussianSplatting2023} & 21.71 & 0.8304 & 0.2548 & 22.69 & 0.8017 & 0.2994 & 18.42 & 0.7170 & 0.3153 & 20.29 & 0.7816 & 0.2920 \\

TiNeuVox~\cite{fangFastDynamicRadiance2022} & 19.71 & 0.8109 & 0.3643 & 21.26 & 0.8289 & 0.3967 & \cellcolor{zbbest}20.66 & \cellcolor{zbthird}0.8145 & 0.2690 & 21.61 & 0.8234 & 0.2766 \\

HyperNeRF~\cite{parkHyperNeRFHigherdimensionalRepresentation2021a} & 24.59 & 0.8770 & \cellcolor{zbthird}0.1650 & \cellcolor{zbthird}25.58 & \cellcolor{zbbest}0.8949 & \cellcolor{zbsecond}0.1777 & \cellcolor{zbsecond}20.41 & \cellcolor{zbbest}0.8199 & \cellcolor{zbthird}0.1911 & 23.45 & 0.8488 & 0.1990 \\

NeRF-DS~\cite{yanNeRFDSNeuralRadiance2023} & \cellcolor{zbbest}24.91 & \cellcolor{zbthird}0.8741 & \cellcolor{zbbest}0.1737 & 25.13 & 0.8778 & \cellcolor{zbbest}0.1741 & \cellcolor{zbthird}19.96 & \cellcolor{zbsecond}0.8166 & \cellcolor{zbbest}0.1855 & \cellcolor{zbthird}23.60 & \cellcolor{zbthird}0.8494 & \cellcolor{zbthird}0.1816 \\

Deformable-GS~\cite{yangDeformable3DGaussians2023a} & \cellcolor{zbsecond}24.86 & \cellcolor{zbsecond}0.8908 & \cellcolor{zbbest}0.1532 & \cellcolor{zbsecond}26.31 & \cellcolor{zbthird}0.8842 & 0.1783 & 19.67 & 0.7934 & \cellcolor{zbsecond}0.1901 & \cellcolor{zbsecond}24.11 & \cellcolor{zbsecond}0.8524 & \cellcolor{zbbest}0.1769 \\

Ours & \cellcolor{zbthird}24.65 & \cellcolor{zbbest}0.8915 & \cellcolor{zbsecond}0.1561 & \cellcolor{zbbest}27.09 & \cellcolor{zbsecond}0.8880 & \cellcolor{zbthird}0.1782 & 19.75 & 0.7943 & 0.1918 & \cellcolor{zbbest}24.26 & \cellcolor{zbbest}0.8544 & \cellcolor{zbsecond}0.1801 \\

\hline
\end{tabular}}
\vspace{-5pt}
\caption{\textbf{Quantitative comparison on NeRF-DS~\cite{yanNeRFDSNeuralRadiance2023} dataset per-scene}. We color each cell as \colorbox{zbbest}{best}, \colorbox{zbsecond}{second best}, and \colorbox{zbthird}{third best}. Our method, overall, achieves the best rendering quality and robust convergence in the majority of scenes. It is worth noting that the metrics we used are the same as those in the main text, with LPIPS using the VGG network.}
\label{tab: nerfds-per}
\end{table*}

The deformation field learns the temporal offsets of Gaussian points. It subsequently adds these offsets to the original values, effectively separating the invariant and variant components of the features. However, as the deformation field learns all Gaussian points, including a substantial number of static points that remain unchanged over time, we introduce a static constraint to diminish the impact of the deformation field on these points. This constraint aids in better distinguishing between static and dynamic points.

Initially, we determine whether a point is static or dynamic by evaluating the magnitude of the coordinate offset $\Delta \mu$ output by the deformation field. If the L1 norm of the point coordinate offset $\|\Delta \mu\|$ is less than a designated $threshold$, which is set at $0.1$, the point is considered as static. Otherwise, it is considered as dynamic. For static points, we compute the static constraint weight for each point based on the reciprocal of the $\|\Delta \mu_i\|$:
\begin{equation}
    \beta_i = \frac{1}{\|\Delta \mu_i\|}, \omega_i = \frac{\beta_i}{\sum_{i=1}^N \beta_i}, \|\Delta \mu_i\| < thresold.
\end{equation}
This is implemented to mitigate the influence of static constraints on dynamic points. Subsequently, we compute the static constraint based on the weight assigned to each point:
\begin{equation}
    \mathcal{L}_s=\sum_{i=1}^N \omega_i \|\Delta\mu_{\mathrm{i}}\|, \| \Delta \mu_i \| < thresold.
\end{equation}
For the constraints on dynamic points, we propose a prior assumption that, at the same timestamp, the majority of dynamic points should exhibit similar motion trends, although there may be some points with different motion trends. Building on this prior assumption, we decompose the coordinate offsets $\Delta \mu$ of all Gaussian points into motion along the three coordinate axes, $\Delta x, \Delta y, \Delta z$. For the motion along each coordinate axis, we further segregate it into positive and negative directions, such as $\Delta x < 0$ and $\Delta x > 0$. Hereafter, we illustrate the consistency constraint of motion using the instance of Gaussian points moving along the positive direction of the $x$ axis:
\begin{equation}
    \mathcal{L}_{con_x^{+}}  =\frac1N\sum_{i=1}^N \| \Delta x_i- \overline {\Delta x} \|,\mathrm{where~}  \Delta x_i >0.
\end{equation}
where  $\overline{\Delta x}$ denotes the average of $\Delta x_i$ where $\Delta x_i > 0$. The objective of this constraint is to diminish the variance of Gaussian points moving along the positive direction of the $x$ axis in their offsets, thus promoting more consistent motion. Similarly, we also compute $\mathcal{L}_{con_x^{-}}, \mathcal{L}_{con_y^{+}}, \mathcal{L}_{con_y^{-}}, \mathcal{L}_{con_z^{+}}, \mathcal{L}_{con_z^{-}}$, which represent the consistency constraints for both positive and negative directions along each coordinate axis.
\begin{equation}
    \mathcal{L}_{con} =\mathcal{L}_{con_x^{+}}+\mathcal{L}_{con_x^{-}}+\mathcal{L}_{con_y^{+}}+\mathcal{L}_{con_y^{-}}+\mathcal{L}_{con_z^{+}}+\mathcal{L}_{con_z^{-}}.
\end{equation}
While this method may appear simplistic, it aptly distinguishes the diverse motion trends of each point and accomplishes the loss constraint for dynamic consistency.

\begin{figure*}
    \centering
    \includegraphics[width=\linewidth]{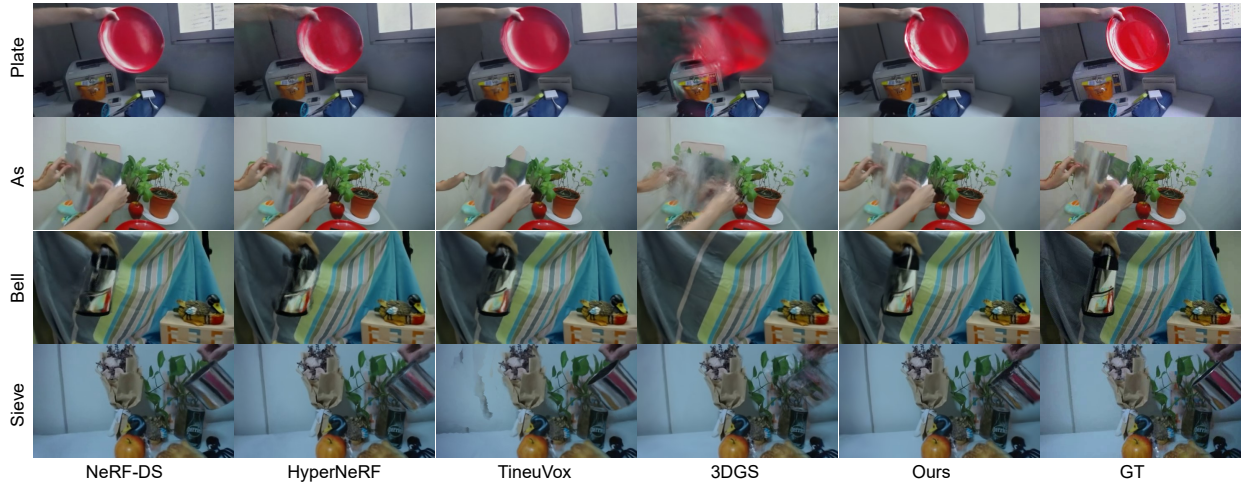}
    \caption{\textbf{Qualitative comparisons of baselines and our method on NeRF-DS real-world dataset.} Experimental results have indicated our ability to mitigate certain high-frequency errors through the utilization of hash coding and denoising masks.}
    \label{fig:NeRF-DS}
\end{figure*}

\begin{figure*}
    \centering
    \includegraphics[width=\linewidth]{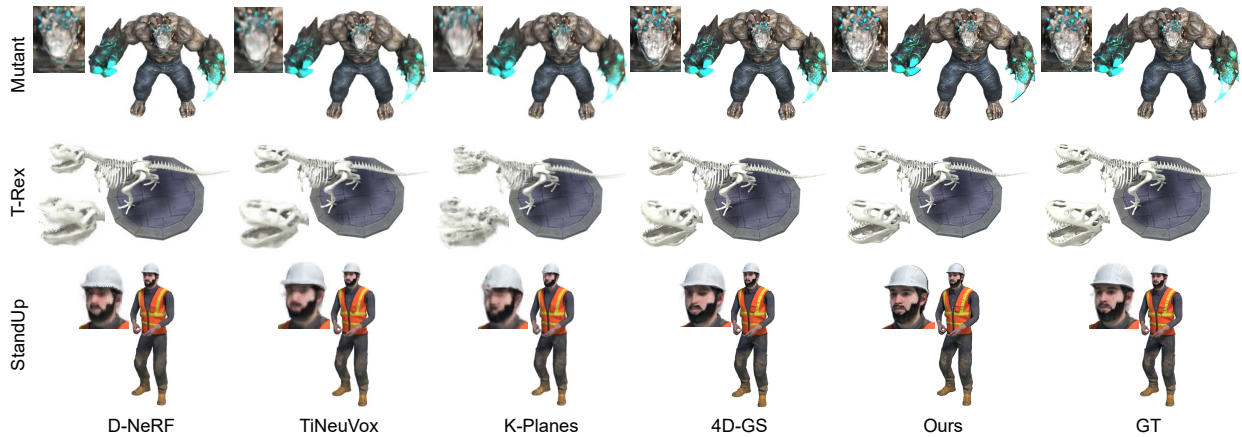}
    \caption{\textbf{Qualitative comparisons of baselines and our method on monocular synthetic dataset.} The experiments demonstrate the efficacy of leveraging deformation fields and hash encoding to facilitate dynamic scene rendering with 3D Gaussians.}
    \label{fig:D-NeRF}
\end{figure*}

\begin{table}[]
\centering
\begin{tabular}{l||lll}
\hline
        Method & PSNR $\uparrow$ & SSIM $\uparrow$ & LPIPS $\downarrow$ \\ \hline
        3D-GS~\cite{kerbl3DGaussianSplatting2023} & 23.23 & 0.9309 & 0.0747 \\
        D-NeRF~\cite{pumarolaDNeRFNeuralRadiance2021} & 30.44 & 0.9649 & \cellcolor{zbthird}0.0508 \\
        TiNeuVox~\cite{fangFastDynamicRadiance2022} & \cellcolor{zbthird}31.97 & 0.9657 & 0.0542\\ 
        Tensor4D~\cite{shaoTensor4DEfficientNeural2023} & 26.91 & 0.9391 & 0.0588 \\ 
        K-Planes~\cite{fridovich-keilKPlanesExplicitRadiance2023} & 31.10 & \cellcolor{zbthird}0.9698 & 0.0452 \\ 
        4D-GS~\cite{wu4DGaussianSplatting2023} & \cellcolor{zbsecond}33.36 & \cellcolor{zbsecond}0.9867 & \cellcolor{zbsecond}0.0266 \\ \hline
        Ours & \cellcolor{zbbest}38.02 & \cellcolor{zbbest}0.9871 & \cellcolor{zbbest}0.0175 \\ 
        \hline
\end{tabular}
\caption{\textbf{Metrics on D-NeRF dataset}. We computed the mean of the metrics across all eight scenes. Cells are highlighted as follows: \colorbox{zbbest}{best}, \colorbox{zbsecond}{second best}, and \colorbox{zbthird}{third best}.}
\label{exp:dnerf-quant}
\end{table}

\section{Experiment}

\subsection{Datasets and Evaluation Metrics}

\par In this section, to validate our method's effectiveness, we tested it on three datasets: the synthetic dataset D-NeRF~\cite{pumarolaDNeRFNeuralRadiance2021}, and two real-world datasets, HyperNeRF~\cite{parkHyperNeRFHigherdimensionalRepresentation2021a} and NeRF-DS~\cite{yanNeRFDSNeuralRadiance2023}. The image resolutions employed in all experiments remain consistent with those specified in the original papers to ensure a fair comparison. 
The metrics utilized for evaluating performance include Peak Signal-to-Noise Ratio (PSNR), Structural Similarity (SSIM), and Learned Perceptual Image Patch Similarity (LPIPS)~\cite{zhangUnreasonableEffectivenessDeep2018}. 
In addition, we conducted a comparison of rendering efficiency, including metrics such as frames per second (FPS) and storage usage.

\subsection{Experiment Details}

\par We implemented our framework using PyTorch. Throughout the training process, we executed a total of 40,000 iterations. During the initial 1,500 iterations, our focus was on training 3D Gaussians to attain stability. Subsequently, we embarked on joint training involving the 3D Gaussians, the deformation hash field, and the denoising mask. The deformable MLP structure maintains consistency with Deformable-GS~\cite{yangDeformable3DGaussians2023a}. The hash table size is set to $2^{20}$, while the resolution ranges from 16 to 2048, and the dimensionality of each layer of features is fixed at 2. The structure of tiny MLP is $64 \times 2$, and the output dimension is 3.

\par Regarding the computation of loss between rendered images and real images, we adhered to the approach outlined in 3D-GS~\cite{kerbl3DGaussianSplatting2023}, utilizing $\mathcal{L}_{D-SSIM}$ and $\mathcal{L}_{1}$. We assigned weights to the loss functions $\mathcal{L}_{dn}$, $\mathcal{L}_s$, $\mathcal{L}_{con}$, and $\mathcal{L}_{m}$, setting them at $1e-2$, $1e-3$, $1e-3$, and $5e-4$, respectively. Notably, $\mathcal{L}_s$ and $\mathcal{L}_{con}$ contributed to training after 3,000 iterations, while $\mathcal{L}_{dn}$ joined after 5,000 iterations.

\par For optimization, we employed the Adam optimizer, with distinct learning rates assigned to different components: the 3D Gaussians retained a consistent learning rate as per the official implementation, while the learning rate for the deformation field decayed exponentially from $8e-4$ to $1.6e-6$, and those for the hash encoding and tiny MLP similarly decayed from $8e-4$ to $3.2e-4$. The $\beta$ values for the Adam optimizer fell within the range of (0.9, 0.999). Experiments conducted on synthetic datasets were executed against a black background at a resolution of 800x800. All experiments were performed on an NVIDIA RTX 3090 GPU.

\begin{table*}[!ht]
    \centering
    \begin{tabular}{ccccccc}
    \hline
        \multirow{2}{*}{\makecell{Denoising \\mask}} & 
        \multirow{2}{*}{\makecell{Static \\constraint}} & 
        \multirow{2}{*}{\makecell{Consistency \\constraint}} & 
        \multirow{2}{*}{\makecell{PSNR $\uparrow$}} & 
        \multirow{2}{*}{\makecell{FPS $\uparrow$}} & 
        \multirow{2}{*}{\makecell{Num $\downarrow$ \\ (k)}} & 
        \multirow{2}{*}{\makecell{Storage $\downarrow$ \\ (MB)}} \\ \\ \hline
        \XSolidBrush & \XSolidBrush  & \XSolidBrush  & 23.67 & 40 & 117 & 7.0 \\ 
        \Checkmark & \XSolidBrush  & \XSolidBrush & 23.95 & 80 & 63 & 3.4 \\ 
        \Checkmark & \Checkmark & \XSolidBrush & 24.10 & 80 & 63 & 3.5 \\ 
        \Checkmark & \Checkmark & \Checkmark & 24.26 & 80 & 63 & 3.5 \\ \hline
    \end{tabular}
    \caption{\textbf{Ablation Study.} We conducted ablation experiments on the NeRF-DS~\cite{yanNeRFDSNeuralRadiance2023} dataset, revealing that the denoising mask effectively eliminates a substantial number of invalid points, thereby enhancing both rendering quality and efficiency. Concurrently, the incorporation of static constraint and dynamic constraint proves effective in constraining the irregular motions of points.} \label{exp:ablation_study}
\end{table*}

\begin{figure}
    \centering
    \includegraphics[width=\linewidth]{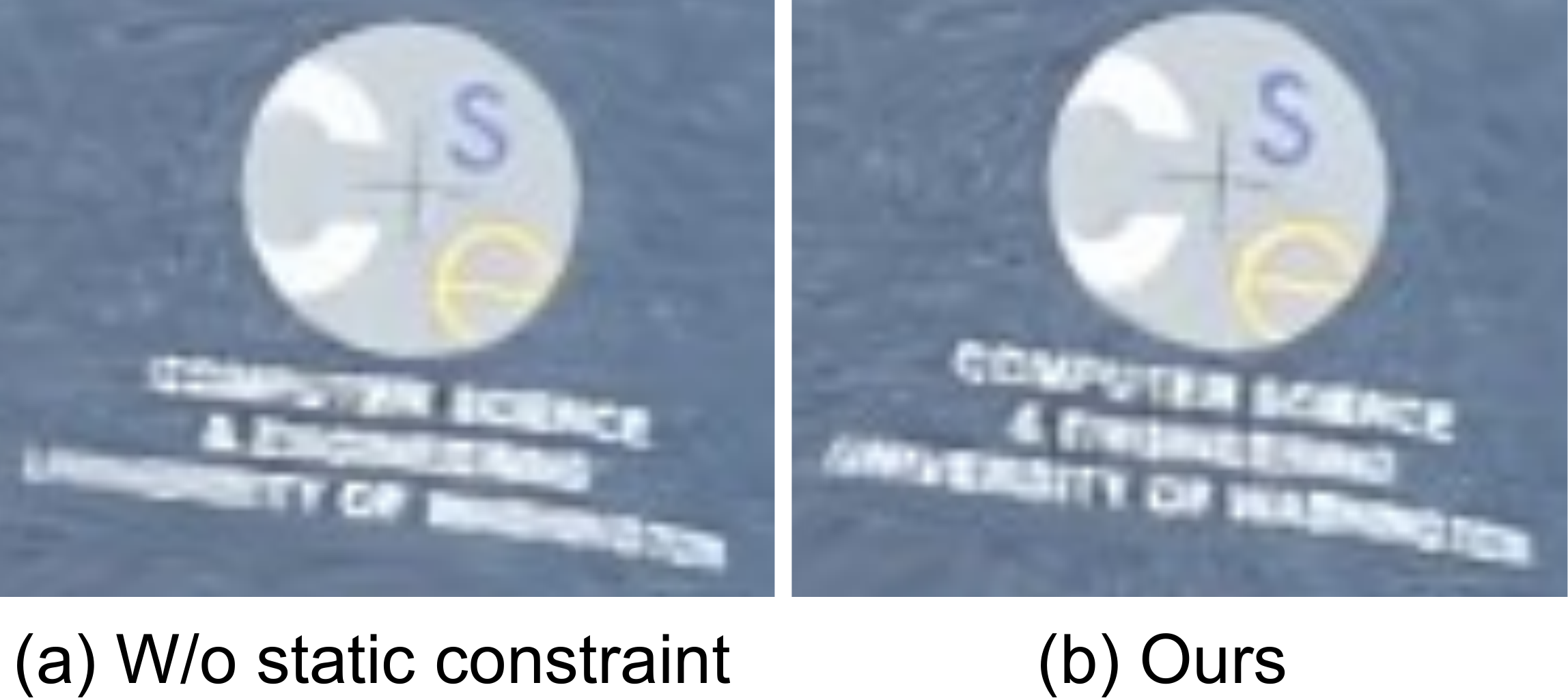}
    \caption{Qualitative comparison of static constraint.}
    \label{fig:ablation_static}
\end{figure}

\subsection{Comparison and Analysis}

\subsubsection{Comparisons on real-world dataset}
We compared our method with previous methods using the monocular real-world datasets NeRF-DS~\cite{yanNeRFDSNeuralRadiance2023} and HyperNeRF~\cite{parkHyperNeRFHigherdimensionalRepresentation2021a}. Table~\ref{tab: nerfds-per} presents the results of quantitative evaluation on the NeRF-DS dataset, indicating the effectiveness of our method, particularly in maintaining structural consistency in dynamic scenes, as evidenced by metrics such as PSNR and SSIM.

\par For a more intuitive evaluation, we provide qualitative results in Figure~\ref{fig:NeRF-DS}. These results demonstrate that our method maintains good performance even in situations where camera poses are not entirely accurate. Given inaccuracies in certain camera poses within the HyperNeRF dataset, we selected only a subset of scenes to verify the robustness of our method, as depicted in Figure~\ref{fig:ablation_static}.

\subsubsection{Comparisons on synthetic dataset} 
We compared our method using the monocular synthetic dataset introduced by D-NeRF~\cite{pumarolaDNeRFNeuralRadiance2021}. Quantitative evaluation results are illustrated in Figure~\ref{fig:D-NeRF}, while qualitative evaluation results are presented in Table~\ref{exp:dnerf-quant}. Notably, our method showcases substantial advantages over NeRF-based approaches on this dataset. Furthermore, it notably outperforms 4D-GS, an extension of 3D-GS, in terms of performance superiority.

\begin{table}[]
\centering
\resizebox{0.38\textwidth}{!}{
\begin{tabular}{l || ccc}
\hline
        Method & FPS $\uparrow$ & \multirow{2}{*}{\makecell{Num $\downarrow$ \\ (k)}} & \multirow{2}{*}{\makecell{Storage $\downarrow$ \\ (MB)}} \\ \\ \hline
        3D-GS & \cellcolor{zbbest}339 & 364 & 86.7 \\ 
        Deformable-GS & 34 & 236 & 56.0 \\ \hline
        Ours & 81 & \cellcolor{zbbest}63 & \cellcolor{zbbest}3.5 \\ \hline
\end{tabular}}
\caption{\textbf{FPS and Storage on NeRF-DS dataset}. We computed the mean of the metrics across all seven scenes. Cells are highlighted as follows: \colorbox{zbbest}{best}.}
\label{exp:nerfds-fps}     
\end{table}

\begin{figure}
    \centering
    \includegraphics[width=\linewidth]{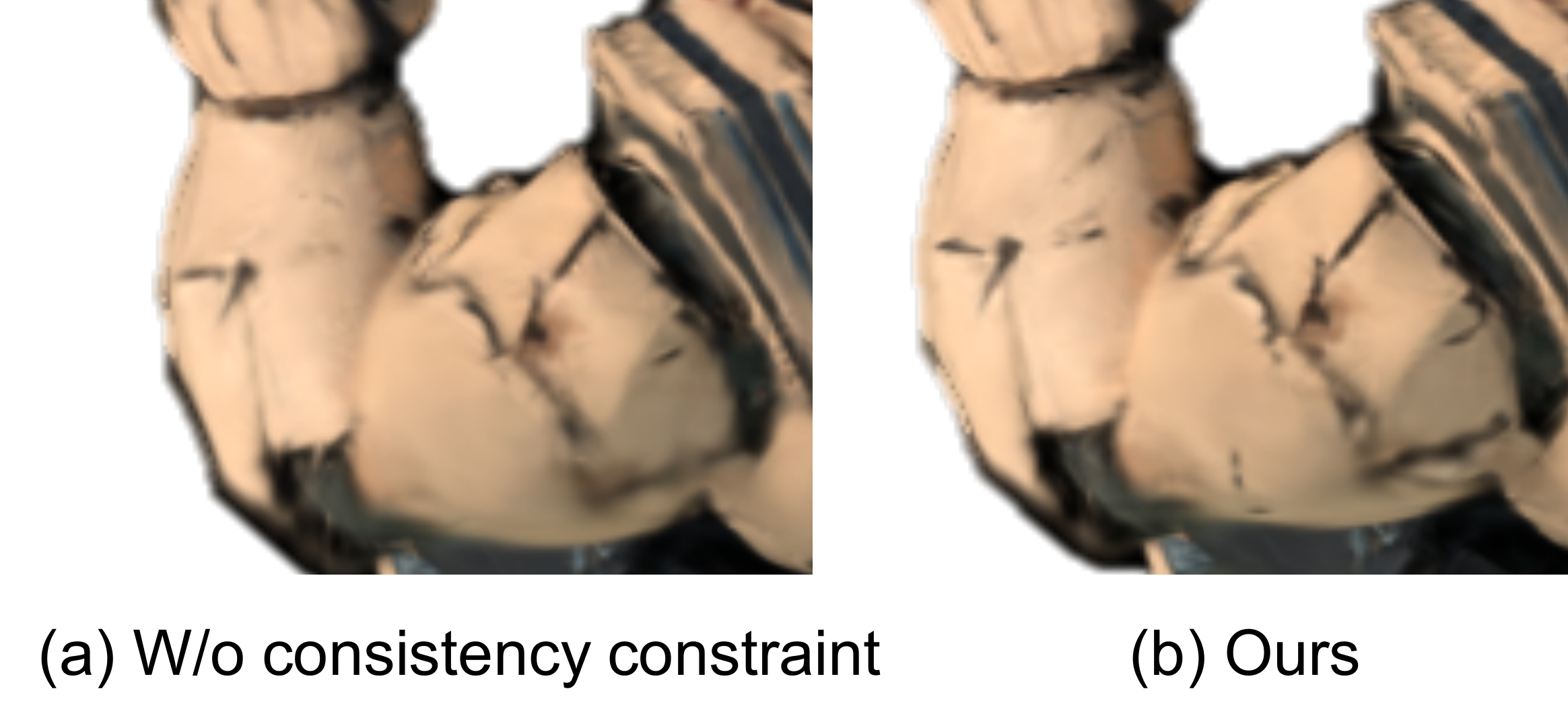}
    \caption{Qualitative comparison of static constraint.}
    \label{fig:ablation_consistency}
\end{figure}

\subsection{Ablation Study} 
Table~\ref{exp:ablation_study} demonstrates the efficacy of the denoising mask in removing unnecessary Gaussian points while improving rendering accuracy. Moreover, the static constraint restricts the impact of static points on deformation field training, depicted in Figure~\ref{fig:ablation_static}. The consistency constraint ensures the motion consistency of the point cloud at a certain timestamp, as illustrated in Figure~\ref{fig:ablation_consistency}.

\begin{figure}
    \centering
    \includegraphics[width=\linewidth]{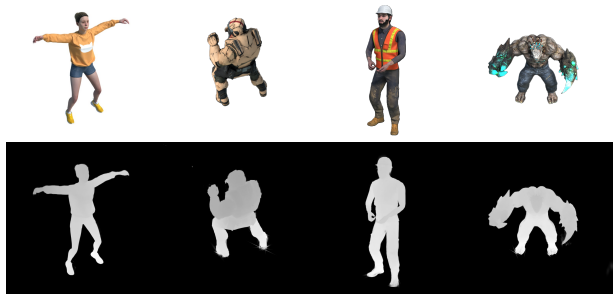}
    \caption{\textbf{Depth Visualization.} We visualized the depth images of the D-NeRF dataset.}
    \label{fig:D-NeRF_depth}
\end{figure}

\subsection{Rendering Effciency} 
Rendering speed correlates directly with the number of 3D Gaussian points. Throughout rendering, we precompute the hash encoding and color derived from the the tiny MLP. Comparative analysis against Deformable-GS~\cite{yangDeformable3DGaussians2023a} reveals our capacity to render dynamic scenes using fewer Gaussian points, as evidenced in Table~\ref{exp:nerfds-fps}.

\subsection{Depth Visualization} 
We showcase our method's proficiency in rendering depth images on the D-NeRF dataset, as depicted in Figure~\ref{fig:D-NeRF_depth}. The precise depth representation underscores the accuracy of our geometric reconstruction, offering significant advantages for tasks involving novel viewpoint synthesis.

\section{Conclusion}

This paper proposes a novel dynamic scene rendering framework that integrates hash encoding, deformation fields, and 3D Gaussians, alongside the introduction of denoising masks and motion consistency constraints to effectively eliminate scene noise. Experimental results showcase that our proposed rendering framework notably reduces memory usage while ensuring high-quality rendering and meeting real-time requirements. On the NeRF-DS dataset, our method achieves SOTA performance. However, our work still exhibits some limitations. For instance, the combination of hash encoding and tiny MLP may inadequately capture high-frequency color features, potentially resulting in image rendering lacking detail in certain cases. Additionally, inaccuracies in pose estimation within real-world scene datasets may lead to blurring phenomena in rendered images. Addressing these issues will be the focus of future work, aiming to enhance the robustness and applicability of the rendering framework.

\bibliographystyle{unsrt}
\bibliography{main}

\end{document}